# FallDeF5: A Fall Detection Framework Using 5G-based Deep Gated Recurrent Unit Networks

**MABROOK S. AL-RAKHAMI[1], (Member, IEEE), ABDU GUMAEI[1, 2], (Member, IEEE),
MOHAMMAD MEHEDI HASSAN[1], (Senior Member, IEEE), BADER FAHAD ALKHAMEES[3],
KHAN MUHAMMAD[4], (Member, IEEE), AND GIANCARLO FORTINO[5], (Senior Member, IEEE)**

[1] Research Chair of Pervasive and Mobile Computing; Information Systems Department, College of Computer and Information Sciences, King Saud University, Riyadh 11543, Saudi Arabia.
[2] Computer Science Department, Faculty of Applied Sciences, Taiz University, Taiz 6803, Yemen.
[3] Information Systems Department, College of Computer and Information Sciences, King Saud University, Riyadh 11543, Saudi Arabia.
[4] Department of Software, Sejong University, Seoul 143-747, South Korea
[5] DIMES, University of Calabria and ICAR-CNR, Rende (CS), Italy

Corresponding author: Mohammad Mehedi Hassan (e-mail: mmhassan@ksu.edu.sa)

The authors are grateful to the Deanship of Scientific Research, king Saud University for funding through Vice Deanship of Scientific Research Chairs.

**ABSTRACT** Fall prevalence is high among elderly people, which is challenging due to the severe consequences of falling. This is why rapid assistance is a critical task. Ambient assisted living (AAL) uses recent technologies such as 5G networks and the internet of medical things (IoMT) to address this research area. Edge computing can reduce the cost of cloud communication, including high latency and bandwidth use, by moving conventional healthcare services and applications closer to end-users. Artificial intelligence (AI) techniques such as deep learning (DL) have been used recently for automatic fall detection, as well as supporting healthcare services. However, DL requires a vast amount of data and substantial processing power to improve its performance for the IoMT linked to the traditional edge computing environment. This research proposes an effective fall detection framework based on DL algorithms and mobile edge computing (MEC) within 5G wireless networks, the aim being to empower IoMT-based healthcare applications. We also propose the use of a deep gated recurrent unit (DGRU) neural network to improve the accuracy of existing DL-based fall detection methods. DGRU has the advantage of dealing with time-series IoMT data, and it can reduce the number of parameters and avoid the vanishing gradient problem. The experimental results on two public datasets show that the DGRU model of the proposed framework achieves higher accuracy rates compared to the current related works on the same datasets.

**INDEX TERMS** 5G, Deep learning, Edge computing, Fall detection, Healthcare system, Internet of Medical Things

## I. INTRODUCTION

Progress in medicine and healthcare has significantly increased average life expectancy, which exceeds the age of 80 years [1]. It is expected that by 2060, the demographic elderly-age dependency percentage will rise [2]. The European Union expects an increase from approximately 28% to 50%, while Asia is expecting an increase from 33% to 45%. Due to such changes, the percentage of elderly people demanding additional help is also predicted to increase [2]. A large number of studies have been conducted in the field of ambient assisted living systems (AALSs) based on the intelligent internet of things (IoT) [3]. Accordingly, the elderly are promised personal health tracking and monitoring on a higher level. For now, sports and fitness are the main leaders in developing IoT-based health-tracking apps [4]. Abnormal event detection is still a growing research field, which exploits the use of smartphones, sensors, and vision-based surveillance cameras [5]. Accident and disease monitoring, together with preventive healthcare services, are also catching up and becoming more common [6]. Due to today's aging population, real-time preventive healthcare applications such as fall detectors are growing in popularity. Need among the elderly is rising for timely and effective help, especially in the event of falls.

Falling is a major cause of serious injury that leads to death among the elderly [7]. Fall detection seeks to automate the alerts and notification system and provide assistance whenever an accident such as a fall occurs [8]. Some detectors provide wearable devices to monitor the user's biomedical data in real-time [9]. This is a notable area because both the number and quality of sensors in smartphones and tracking devices are continually growing. The collected data is increasingly accurate, easier to monitor, and serves as a great base for creating an efficient system for fall detection [10].

The primary purposes of fall detection apps are automatic fall detection and alerting the assigned caretaker. The geriatric care field will benefit significantly from the availability of fall detection systems due to the greater prevalence of falls, as well as falls with severe consequences, among elderly people [11]. It is essential to provide patients, especially the elderly, with a non-invasive tracking system that is easy to use. A heavy and complicated wearable sensor might not be the best choice option [1]. In addition, it is crucial to minimize the device's power usage, enabling patients to use it longer, even when they forget to recharge. Optimizing the energy usage of the system and device is vital in elderly health tracking.

Prior studies on fall detection have focused on specially designed hardware and software, most of which require a high cost [12]. In [12], the authors reviewed 57 projects in which wearable fall detection devices were analyzed. Only 7.1% of the projects were tested in real-world conditions. The authors also demonstrated the utility of a fall detection wristwatch device. Not only are they non-intrusive but also they are unlikely to cause injuries when an accident occurs. In comparison, sensors worn around the torso were associated with greater fall detection accuracy [13], but they were less likely to receive a positive review from the elderly users due to their being less comfortable and causing more distress.

Compared to smartphone-based fall detection solutions [14-16], smartwatches are preferred in our research, especially for elderly users. There are two main reasons for this, which are stated as follows:
- Smartphone Placement and Usability Issues: Smartwatches are easier to wear and lighter in weight, and thus are a suitable option for elderly users.
- Energy Consumption and Battery Life: Smartwatch batteries last longer, which is suitable for the elderly, especially if they forget to charge their smartphones.

The main contributions of our work are as follow:
1) We propose a fall detection framework using a 5G-based deep gated recurrent unit (DGRU) neural network and smartwatches as the internet of medical things (IoMT) sensors for older users and patients. This framework is abbreviated as FallDeF5.
2) We propose a framework based on deep learning (DL) for fall detection over mobile edge computing (MEC) in 5G networks, empowering the IoMT.
3) We address the imbalanced data problem in data samples of fall and non-fall labels collected from smartwatch sensors.
4) We use DGRU to classify fall and non-fall cases for automatic and effective fall detection with high accuracy.

Using the DGRU neural network, it is possible to reduce the number of parameters of other variations of recurrent neural networks (RNNs) and avoid the vanishing gradient problem.
5) We fine-tune the parameters of GRU using our DL experience and a grid search tuning algorithm.
6) We evaluate the proposed framework's DL model on two publicly available datasets of smartwatches to validate the applicability of artificial intelligence (AI) in benefitting from IoMT device adoption in intelligent healthcare systems.

The rest of the paper is organized as follows: Section II reviews the tools and technologies used in our work; Section III discusses related works; Section IV presents the proposed framework; Section V describes the experimental results; and finally, Section VI concludes the work.

## II. TOOLS AND TECHNOLOGIES

Due to technological improvement and development, it is possible to integrate and apply multiple innovations to devise increasingly smart applications. In this section, we explain some of the technologies used in our research, which include edge computing, DGRU, Docker containers, and 5G networks.

### A. EDGE COMPUTING

Edge computing is an emerging technology that provides cloud and information technology services with greater proximity to users. Edge computing reduces application latency because its computing and storage capacity is situated at the edge of the network, closer to the end-user. In our previous papers [17, 18], we introduced a new edge computing model that enabled us to reduce latency from processes occurring in the cloud, wherein the data were processed close to the edge. After data collection in the edge, we were able to add another layer, a more generalized structure of the edge, to the cloud. The authors in [19] examined different features of edges, how they can be applied, and how they have improved in our computing paradigm.

Many prior studies have highlighted specific applications of edge computing, including IoT, ad-hoc or vehicular networks [20], and human activity recognition [20]. Some of the benefits of using edge computing in relation to low latency and remote processing are lower bandwidth utility, assured quality of service (QoS), quality of experience (QoE), and real-time notifications. Edge computing has become particularly useful in healthcare due to these advantages. Certain illnesses where the usage and benefits of edge computing are evident include Parkinson's disease [21], ECG and EEG feature extraction [22], and chronic obstructive pulmonary disease [23].

### B. GATED RECURRENT NEURAL NETWORKS (GRU) AND DEEP GATED RECURRENT NETWORKS (DGRU)

To address the vanishing gradient and exploding gradient problems that have influenced back-propagation through time in early attempts at RNNs, we used gated recurrent neural networks (GRU) [24]. These networks are RNN structures that provide a memory cell with nonlinear gating units [25]. This structure of a basic RNN contains only two gates: the update gate and the reset gate.

Another type of RNN is known as a deep gated recurrent unit (DGRU), which is designed for classification tasks. A major difference that marks DGRUs is that they contain $n$ layers. Additionally, it is able to solve GRU issues. Some of these issues are that the GRU process might drop dramatically if it is continuously trained, which happens due to the nature of the softmax activation function and the shallowness of the gate. To control the gate with a more complicated structure, an extra weight matrix is added to $Z_{out}$, which increases network stability. Figure 1 shows the structure of DGRU, where one weight matrix is added before $Z_{out}$ to the previous output.

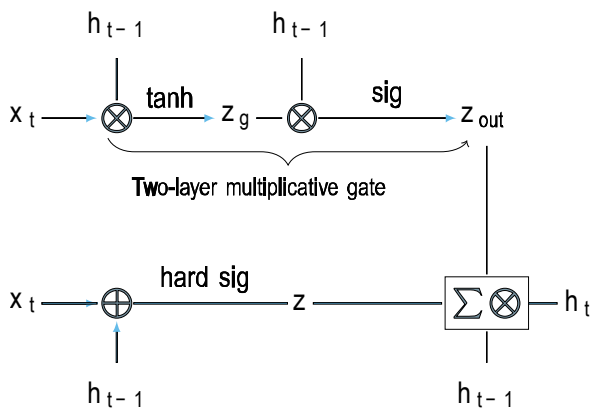

**FIGURE 1.** Structure of DGRU

Based on our experience in deep learning and machine learning methods, as well as trial and error techniques, we built the proposed DGRU model with 2 hidden layers with 64 hidden dimensions and 2 dimensions for the output layer. The batch size and learning rate of the model were initialized at 128 and 0.001, respectively. The number of epochs to train the model was set as 100.

### C. DOCKER CONTAINERS

Organizing the application and its self-containerized dependencies are what make the container so advantageous in the process of Docker usage [26]. A Docker compared to a virtual machine (VM) shows its lightweight component abilities, which significantly assist algorithmic processing. Just like with a VM, the Dockers are isolated, but because the container works without any guest OS, we have a lightweight and easy-to-manage software that runs on the edge device [18]. Figure 2 shows the Docker daemon that manages the creation of containers and images created for the application [18].

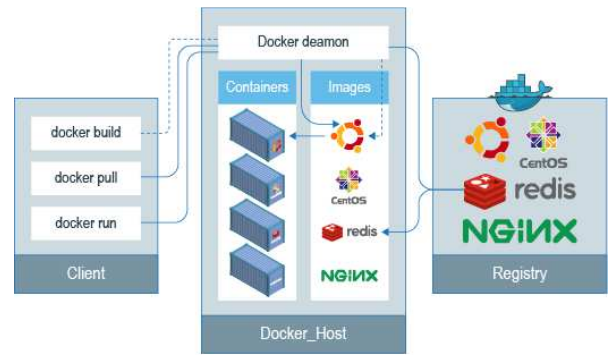

**FIGURE 2.** Docker daemon

Communication and information flow in Docker are possible through the use of a REST API. Docker client commands are processed within the daemon, allowing users to interact with the system. The images created from the Docker file are read-only and are stored in the registry. They are available for use at any time (whenever a pull command is carried out). To run an image, container creation takes place and the application uses the resources contained in the image. Containers work independently and host the OS's kernel, and they are more efficient compared to VMs [27]. In particular, containers function more effectively in offering simple and fast configuration processes, as well as giving minor overhead by isolation. The quality of the isolation also makes it look as if the containers are running each on separate OSs with an assigned memory and central processing unit (CPU).

In our work, we focus on the use of containerization, which increased the speed of the model and improved the maintenance of the app modules in terms of easy navigability. Both detection and decision-making take place in the edge, benefiting in terms of low latency and allowing for near-edge processing. In Section IV, we present more information on the structure of our framework, together with the benefits of using edge computing, deep learning, containerization technology, and 5G networks. In the next sections, we provide a more detailed description of the proposed framework.

### D. 5G TECHNOLOGY

5G networks are being prepared to serve a range of users both for the corporate world and for domestic users. This has resulted in a widespread desire for rapid speeds, security, reliability, and low latency [28]. This is the expectation that 5G technology promises, and it is positioned as the true broadband for lossless use in any handling situation. With its substantial capacity, 5G will be a strong point in the wide usability of the IoT associated with the new IPV6 model [29]. 5G networks will transform the world; their potential is enormous because their applications are not only concerned with mobile services but also are linked to delivery capacity with quick responses. In this way, the population will have services contracted from this technology to meet their needs from a fixed or mobile device. It will meet the needs of government and business projects in different areas [30].

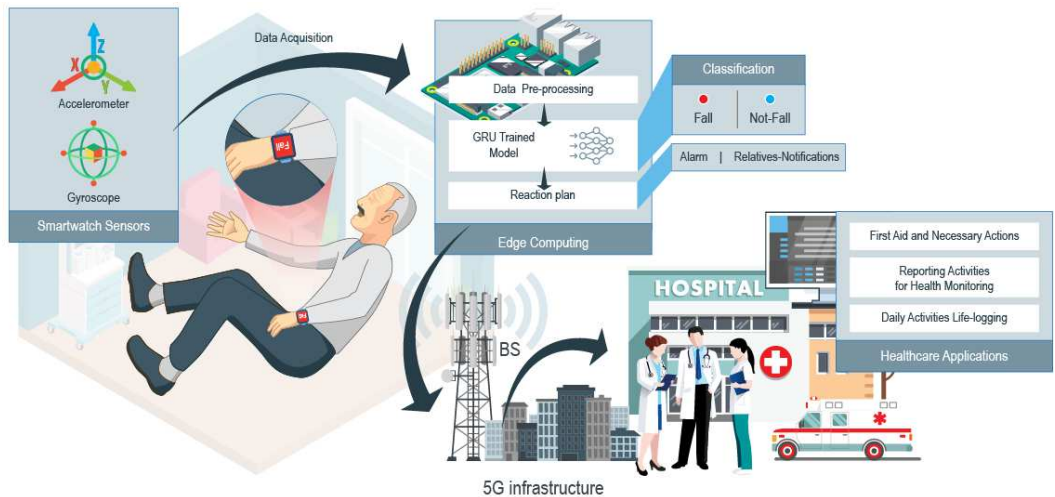

**FIGURE 3.** Proposed fall detection framework based on 5G networks and edge computing.

Generally, the most distinguished features of 5G technology are as follow:
- Incredible speed and super-reliable, real-time connections: Users can achieve ultra-high data rates with instant connections, low latency, and high reliability. This aspect is crucial for, for example, applications based on the exchange of multiple high-quality multimedia streams.
- Optimal service quality even in crowded places: In the future, even in crowded places such as stadiums or concert halls (where connectivity is now heavily compromised by the coexistence of devices that try to access network resources in a concurrent way), a satisfactory browsing experience will be guaranteed.
- Optimal experience on the move: Even users who move at non-negligible speeds (e.g., by car or public transport) will be guaranteed optimal quality of service.

### III. RELATED WORK

It is possible to conduct fall detection research using different technologies and methods. One of the methods is to analyze data from external cameras [31], floor sensors [32], or body-worn sensors such as gyroscopes, accelerometers, and air pressure sensors [33, 34]. For each method, the system collects a stream of data based on the sensors; this allows for a fall to be recognized and distinguished from a non-fall. The collected data are often used to create and train machine learning algorithms, but there are also studies that rely on specific rules (instead of datasets) to differentiate between a fall and a non-fall. Classification of such datasets can show the difference in the quality of a fall detection algorithm. Many studies have developed smartwatch-based fall detection systems in their research [35-37]. However, combining smartwatches with additional sources of information (e.g., sensors not placed on the wrist) is rarely seen in related research.

As mentioned before, most smartwatches and smart bands are viewed as extensions of smartphones [38]. There are many examples of using a wrist sensor together with a smartphone [38, 39]. Sometimes, the sensor connected to the smartphone is used by the detection algorithm. The phone can also serve as the focal point of processing the data from the wrist sensor, which is usually connected via Bluetooth.

The use of such technology is shown in [40], where the authors used an Android smartphone and a Bluetooth-enabled Pebble smartwatch. The watch's energy consumption was reduced to 5 Hz by reducing the sampling rate of the accelerometer. To reduce the usage of wireless connections, the following inactive measurements were also grouped and scheduled to be sent every two seconds. Thanks to these adjustments and the extension of the battery's lifespan, the smartwatch was able to track for 30h continuously.

In [41], the authors deployed an FDS centered around a commodity-based smartwatch paired with an Android smartphone via Bluetooth. Together with the cloud persistence storage and data analysis tools, the detector was set within the Android IoT platform. The Weka Java package was used to implement the support vector machine (SVM) algorithm. Battery consumption of the wrist sensor was not considered in this study. Casilari and Oviedo [42] also decided to use the smartwatch-smartphone combination in their study. The first results indicated that using FDS can significantly lower the smartwatch's battery lifespan. More than 50% of the battery energy (in LG Watch [W110G - R model]) was used after working for 7, making the smartphone the "energy bottleneck" of the FDS. Similarly, Deutsch et al. [43] analyzed a hybrid combination consisting of a Pebble smartwatch and an Android phone connected via Bluetooth. The phone battery died after 17-19 hours of data transmission between the devices.

An Android app, SmartFall, was devised by the authors in [4], which uses accelerometer data obtained from a commodity-based smartwatch IoT to detect falls. The phone connects to the watch through the app, which then calculates the fall predictions in real-time. This avoids the delay in communicating with a cloud server while still maintaining data privacy. The authors applied both traditional (e.g., SVM and naive Bayes) and non-traditional (e.g., DL) machine learning algorithms to build fall detection models with three separate datasets: Smartwatch, Farseeing, and Notch. They concluded that the DL model significantly outperformed the traditional models across the three datasets. This occurs because the DL model is able to acquire the subtle features from the raw accelerometer data automatically, whereas the naive Bayes and SVM models can only learn manually from a small set of extracted features. One of the most important qualities of the DL model is that it shows better performance in helping new users to understand fall prediction. The study also uses a three-layer open IoT system architecture in the SmartFall app, which enables the smooth collection and processing of sensor data modalities (e.g., temperature, heart rate, walking patterns, and skin), in turn offering insights into the user's well-being.

An experimental study of ensemble DL was undertaken to analyze time series data on IoT devices. In previous studies, the researchers concluded that DL outperformed traditional learning methods in fall detection apps that need processing, which was attributed to its ability to acquire the necessary features in time series data automatically.

However, DL networks generally require large datasets for training. Keeping that in mind, it is important to note that there are no large datasets for real-time smartwatch-based fall detection that are publicly accessible in the healthcare domain. Fall data is also quite noisy, and other sounds reaching the wrist sensor may be incorrectly interpreted as a fall. This study focused on applying the combination of DL (in this case, RNNs) and a set of techniques (AdaBoosting and Stacking) to a fall detection app as a case study. Experiments in training different set models were carried out using two separate simulated fall datasets. The authors concluded that a set of DL models, together with the stacking set technique, outperformed a single DL model trained on the selected data samples, making it more appropriate for smaller datasets.

## IV. PROPOSED FRAMEWORK

Figure 3 shows the proposed framework, which focuses on fall detection based on different components. In this section, we described different framework components which are the edge computing, healthcare applications and 5G network. We described the methodology of our work through the framework components.

### A. EDGE COMPUTING COMPONENT

The first layer is called the edge. This part of the structure focuses on real-time data processing and decision-making in a fall scenario. The edge computing component is connected to the smartwatch to receive gyroscope and acceleration sensor signals. The main purpose of this component is to collect movement data from elderly people, complete data pre-processing, deploy the DL model in Docker-based devices, and take action in the event of a fall. During data collection, the pre-processing module portions the collected data into several sections (time windows), enabling each section of 40 points to be analyzed. Figure 3 shows the data portioning process in this component. To assist in pre-processing and data storage, Raspberry Pi devices are used. This allows setting the Docker containerization as a light component when compared to virtual machine techniques. The above processes will enable us to set up and prepare our trained model rapidly and efficiently.

Figure 3 illustrates how each of the structural elements connects and interacts between the components. Smartwatch is connected to smartphone through a Bluetooth radio connection. A smartphone app with two running modes was developed. The first mode receives and categorizes data from the smartwatch, which are later transferred to the Raspberry Pi device for processing in the edge. In turn, a DGRU model is deployed with on edge Docker containers.

### B. 5G INFRASTRUCTURE COMPONENT

The main purpose of 5G networks is to act as a bridge between the intelligent edge environment and smart healthcare applications and services. 5G networks play an integral role in accelerating the transfer of data from the edge to healthcare applications. 5G infrastructure enables interconnected devices to distribute data across multiple devices. The IoMT requires high connectivity to transfer and manage data that is transmitted easily and smoothly. Hence, 5G has made this possible as it is possible to transfer patient data (e.g., medication and diagnostic information) with ease across medical systems. Furthermore, the stakeholders in charge, including caregivers and healthcare professionals, can make meaning out of it, saving time and cost and reducing emergency cases. This helps in the provision of high-quality healthcare services.

### C. HEALTHCARE APPLICATIONS COMPONENT

The third component involves different healthcare applications, which could focus on enabling timely reactions to falls. Reports and monitoring dashboards are available for doctors and other healthcare providers. Life-logging daily activates and periodic healthcare checks are also reported in this component.

Edge computing makes patient data readily available for doctors, including information on diagnostics, medication, and other relevant data. Hence, IoMT is being applied extensively in clinical settings for efficiency, effectiveness, and quality improvement. Additionally, the technology is finding broad adoption in the homes of elderly individuals, where sensors are used to collect data on human activities, transmitting it for processing and use. Caregivers are alerted in the event of an emergency (e.g., a fall), which enables a timely response. The fact that more people are connected, as well as the fact that patients are connected to their doctors, means that patient data will be transmitted regularly, and the doctors can utilize it in a timely manner.

Figure 4 illustrates the data pre-processing phase, in which signal values are segmented for training and detection. For model training, data segments were labeled using two classes: Fall and ADL (activities of daily living, i.e., non-fall). This shows how the data signals of human activities were pre-processed easily and transferred on time.

## V. EXPERIMENT

This section describes the experimental study used for the proposed framework, as well as the datasets adopted and the sub-processes applied to evaluate its applicability.

### A. DATASETS DESCRIPTION

In this subsection, we present the datasets used to evaluate the DL model of the proposed framework. We conducted the experiments on two public datasets obtained from smartwatches. These datasets are described below.

#### 1) SMARTWATCH DATASET

The first dataset used in our experiment was the Smartwatch dataset [4], which collected data from 7 volunteers wearing an MS Band watch. All participants were checked to ensure good health as a necessary step to perform simulated falls and ADLs. They settled in the range of 21-55 years old, 5 ft. to 6.5 ft. tall, and weighing from 100 lbs. to 230 lbs. Each volunteer was asked to wear the smartwatch on their left wrist and was then tested during different ADL action sets and movements, including jogging, sitting down, throwing, and waving. This ADL set, labeled as non-fall, was chosen based on the broad usage of arms and the neutrality of the movements. In the second part of the smartwatch dataset collection, the volunteers performed four types of falling. They were asked to simulate a fall onto a 12-inch mattress on different sides: front, back, left, and right. Each type of fall was repeated 10 times. This experiment was conducted with sampling rates of 4 Hz, 1.25 Hz, and 62.5 Hz supported by the smartwatch and settled with 31.25 Hz. The dataset consists of 51,249 data points labeled as either fall or non-fall activities. 34,020 were allocated to the training set and 17,229 to the test set.

#### 2) SMARTFALL DATASET

The second dataset selected for our research was the SmartFall dataset. This dataset was obtained from 14 volunteers wearing an MS Band 2. As in the previous dataset, the volunteers were checked to ensure good health before performing the simulated falls and ADLs. They settled in the range of 21-60 years old, 5 ft. to 6.5 ft. tall, and weighing from 100 lbs. to 230 lbs. In the second part of the smartwatch dataset collection, the volunteers performed four types of falls. They were asked to simulate a fall onto a 12-inch mattress on different sides: front, back, left, and right. Each type of fall was repeated 10 times.

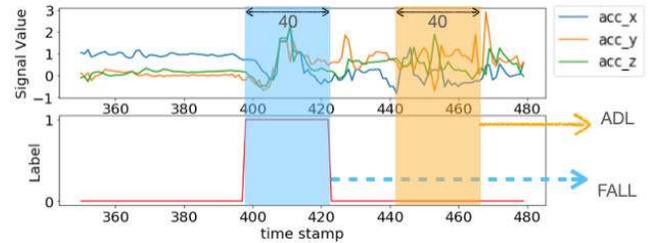

FIGURE 4. Data pre-processing phase

This experiment was carried out based on the sampling rate of 31.25 Hz. Data were collected through a smartphone (Nexus 5X, 1.8 GHz, Hexa-core processors with 2G of RAM), which was connected to the smartwatch. The watch labeled falls as falls whenever the smartwatch button was pressed. Otherwise, the data point was labeled as a non-fall. With a researcher holding the smartphone and observing the volunteer, this enabled us to collect real-time data.

Even with the help of a researcher next to the volunteer, errors could have occurred in data collection. In particular, the button could be pressed too soon, too late, or too long. To reduce the possibility of error, a post-processing step was applied to the collected data, ensuring that the fall label was assigned correctly in relation to the critical fall phase. An R script was applied to the processing, enabling us to check whether the fall label was correctly assigned to the data, the highest peak of speed, and data points before and after that point. The dataset consisted of 183,806 points, each labeled as fall or non-fall activities and divided into a training set (92,781) and test set (91,025).

### B. EVALUATION METRICS

We measured the performance of the proposed framework by calculating accuracy, precision, recall, and F1-score. These metrics can be computed as below:

$$\text{Accuracy} = \frac{\text{TP} + \text{TN}}{\text{TP} + \text{TN} + \text{FP} + \text{FN}} \quad (1)$$

$$\text{Precision} = \frac{\text{TP}}{\text{TP} + \text{FP}} \quad (2)$$

$$\text{Recall} = \frac{\text{TP}}{\text{TP} + \text{FN}} \quad (3)$$

where TP represents the count of true positive instances that were correctly detected; FP is the number of false positive instances that were incorrectly detected; TN represents the count of true negative instances that were correctly detected; and FN represents the count of false negative instances that were incorrectly detected. Precision is the ratio between true positives (number of times that a class is detected correctly) and the sum of true positives and false positives. Recall (also known as sensitivity) refers to the ratio between the number of true positives and the sum of true positives and false negatives.

## C. RESULTS AND DISCUSSION

During the experiments, the training and test data points across the two datasets were processed to generate data instances using a 40-point window size. Moreover, we labeled the generated data instances by assigning a fall label to any instance with 25 signal points of fall labels within the window size. Otherwise, the data instance was labeled as non-fall. Tables I and II show the training and test instances for the two datasets.

TABLE I
INSTANCES DISTRIBUTION IN TRAINING AND TEST SETS OF PROCESSED DATASET 1

| Class Label | Training Set | Test Set | Total |
| --- | --- | --- | --- |
| fall | 2912 | 1456 | 4368 |
| non-fall | 31068 | 15733 | 46801 |
| Total | 33980 | 17189 | 51169 |

TABLE II
INSTANCES DISTRIBUTION IN TRAINING AND TEST SETS OF PROCESSED DATASET 1

| Class Label | Training Set | Test Set | Total |
| --- | --- | --- | --- |
| fall | 5232 | 3216 | 8448 |
| non-fall | 87509 | 87769 | 175278 |
| Total | 92741 | 90985 | 183726 |

As shown in Tables I and II, the numbers of training instances in the two datasets were highly imbalanced. The imbalanced classes of the training sets mean that the DL algorithm is skewed or biased to the majority class rather than the minority class, which leads to poor classification results. To resolve this problem, we downsampled the non-fall data portion to ensure it was the same as the fall data portion. Consequently, we randomly selected 2,912 from 31,068 instances of the non-fall class for the Smartwatch training set and 5,232 from 87,509 instances for the SmartFall training set. Furthermore, in the training phase of the DGRU model, 10% of each training set was used as a validation set. Also, we initialized the DGRU's parameters with random values. After training the model several times with different parameters, as well as after drawing on our experience in deep learning, we set the layers of the model to 2 hidden layers. The learning rate was 0.0001 and the batch size was 128. The hidden units were set to have 256 and the number of epochs was 100. Figures 5 and 7 show the training loss and validation accuracy for the Smartwatch dataset, while Figures 5 and 7 visualize the training loss and validation accuracy of the SmartFall dataset.

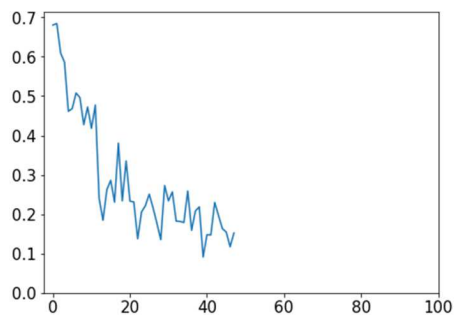

FIGURE 5. Training loss on Smartwatch dataset

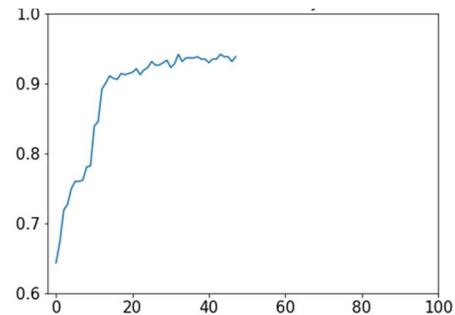

FIGURE 6. Validation accuracy on Smartwatch dataset

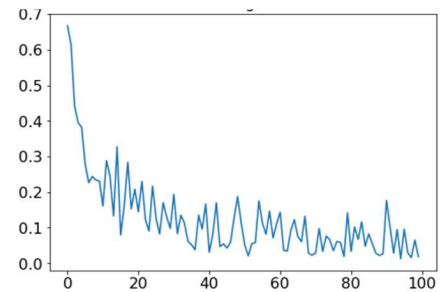

FIGURE 7. Training loss on SmartFall dataset

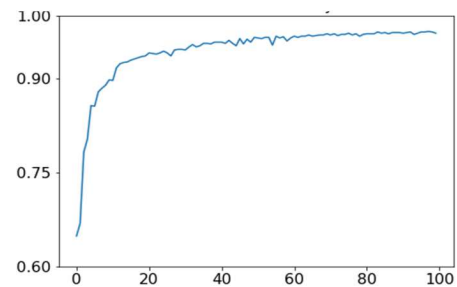

FIGURE 8. Validation accuracy on SmartFall dataset

As shown in Figures 5 and 6, the training process of the model stopped at epoch 48 due to the early stopping mechanism. The early stopping mechanism is a regularization form used to halt training at the epoch after which performance begins to degrade. In contrast to Figures 7 and 8, the training process continued until epoch 100. For the performance of the DGRU model on the Smartwatch dataset, the experimental results of the evaluation metrics (described in Subsection B) are shown in Figure 9 and Table III. Besides, Figure 10 and Table IV show the experimental results of the model on the SmartFall dataset.

To compare our model's performance relative to related recent work, Table V introduces accuracy scores on the two datasets. Clearly, the DGRU model used in the proposed framework achieved high accuracy values compared to other models on the same evaluation datasets.

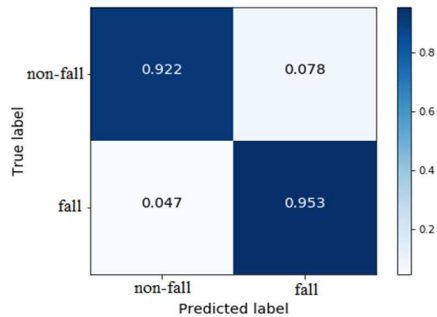

**FIGURE 9.** Confusion matrix on Smartwatch dataset

TABLE III
EVALUATION OF PRECISION AND RECALL OF
TEST SET FOR DATASET 1

| Class Label | Precision | Recall |
|---|---|---|
| non-fall | 0.995 | 0.922 |
| fall | 0.530 | 0.953 |
| Weighted avg. | 0.956 | 0.924 |
| Accuracy | 92.4% | |

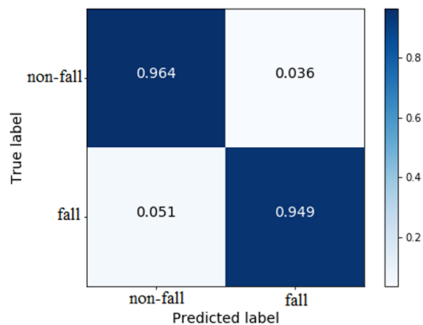

**FIGURE 10.** Confusion matrix on SmartFall dataset

TABLE IV
EVALUATION OF PRECISION AND RECALL OF
TEST SET FOR DATASET 2

| Class Label | Precision | Recall |
|---|---|---|
| non-fall | 0.998 | 0.964 |
| fall | 0.493 | 0.949 |
| Weighted avg. | 0.980 | 0.964 |
| Accuracy | 96.4% | |

It is also noticeable that because the distribution of fall and non-fall classes was imbalanced, the results for the precision coefficient in the fall scenarios were poor. As shown in Table I, the numbers of fall and non-fall instances in the training set were 2912 and 31068, respectively. Also, in Table II, the number of fall instances in the training set was 5,232, while the number of non-fall instances was 87,509.

TABLE V
COMPARISON RESULTS OF PROPOSED FRAMEWORK
AGAINST RELATED WORK

| Ref. (Year) | Dataset | Accuracy (%) |
|---|---|---|
| [4] (2018) | Smartfall | 85 |
| [44] (2019) | Smartwatch | 90 |
| This work (2020) | Smartwatch and Smartfall | 92.4 and 96.4 |

## VI. CONCLUSION

Falling is a major issue for elderly people who often live alone, and it can have severe consequences. Accordingly, timely help is required to ensure the provision of effective first aid and support. Recent advances in artificial intelligence and communications technologies such as 5G can offer substantial help in stimulating innovative, efficient, and effective solutions for the problem of falls. In this work, we proposed a new fall detection framework, which we refer to as FallDeF5. The proposed framework is based on DGRU and the use of MEC in 5G wireless networks to empower IoMT-based healthcare applications. We dealt with time-series IoMT data from smartwatches, and experimental were obtained from two public datasets. The accuracy rates of FallDeF5 compared to the current state of the art were favorable, even on the same datasets. In future work, we will focus on solving the imbalanced class problem using resampling for the training set. Moreover, we will be dedicating our time to enhance the accuracy of FallDeF5 and to conduct experimental studies measuring edge performance.

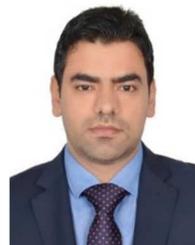

**MABROOK S. AL-RAKHAMI** [M'20] received a master's degree in information systems from King Saud University, Riyadh, Saudi Arabia, where he is currently pursuing a Ph.D. degree with the Information Systems Department, College of Computer and Information Sciences. He has worked as a lecturer and taught many courses, such as programming languages in computer and information science, King Saud University, Muzahimiyah Branch. He has authored several articles in peer-reviewed IEEE/ACM/Springer/Wiley journals and conferences. His research interests include edge intelligence, social networks, cloud computing, Internet of things, big data and health informatics.

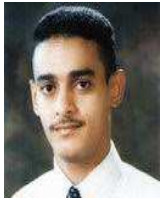
**ABDU GUMAEI** received his Ph.D. degree in computer science from the Computer Science Department at King Saud University in 2019. His main areas of interest are software engineering, image processing, computer vision, and machine learning. He has worked as a lecturer and taught many courses such as programming languages in the Computer Science Department, Taiz University. He has several researches in the field of image processing. He received a patent from the U.S. Patent and Trademark Office in 2013.

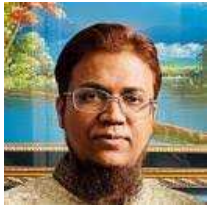
**MOHAMMAD MEHEDI HASSAN** [SM'18] received his Ph.D. degree in computer engineering from Kyung Hee University, Seoul, South Korea, in February 2011. He is currently a professor with the Information Systems Department, College of Computer and Information Sciences, King Saud University, Riyadh, Saudi Arabia. He has authored and coauthored around 210+ publications including refereed IEEE/ACM/Springer/Elsevier journals conference papers, books, and book chapters. His research interests include edge/cloud computing, the Internet of Things, cyber security, deep learning, artificial intelligence, body sensor networks, 5G networks, and social network.

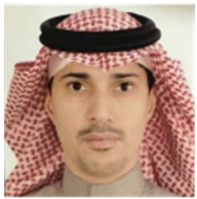
**BADER FAHAD ALKHAMEES** received his Bachelor's degree in Computer Science from King Saud University, Riyadh, Saudi Arabia, in 2003. He received his Master's degree in Software Systems from Heriot Watt University, Scotland, United Kingdom, in 2008. He also received his Ph.D. degree in Biomedical Informatics from Rutgers University, New Jersey, United States. Currently, He is an assistant professor at the Information System Department, College of Computer and Information Sciences, King Saud University. His research interests include Biomedical Informatics, Medical Imaging and Diagnosis, Machine Learning, Fuzzy Systems, Cloud and Edge Computing, the Internet of Things, and Computer Networks.

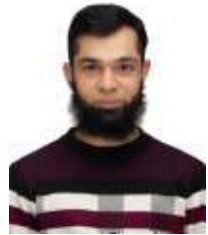
**KHAN MUHAMMAD** received the Ph.D. degree in digital contents from Sejong University, Seoul, South Korea, in 2018. He is currently working as an Assistant Professor with the Department of Software and a Lead Researcher with the Intelligent Media Laboratory, Sejong University. His research interests include intelligent video surveillance (fire/smoke scene analysis, transportation systems, and disaster management), medical image analysis, (brain MRI, diagnostic hysteroscopy, and wireless capsule endoscopy), information security (steganography, encryption, watermarking, and image hashing), video summarization, multimedia, computer vision, the IoT, and smart cities. He has filed/published over seven patents and 120 articles in peer-reviewed journals and conferences in these areas. He is serving as a Reviewer for over 90 well-reputed journals and conferences, from IEEE, ACM, Springer, Elsevier, Wily, SAGE, and Hindawi publishers. He acted as a TPC Member and a Session Chair at more than ten conferences in related areas. He is also an Editorial Board Member of the Journal of Artificial Intelligence and Systems and a Review Editor of the Section "Mathematics of Computation and Data Science" in the journal Frontiers in Applied Mathematics and Statistics.

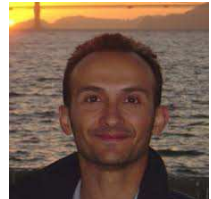
**GIANCARLO FORTINO** [SM'12] is a full professor of computer engineering in the Department of Informatics, Modeling, Electronics, and Systems at Unical. He received a Ph.D. in computer engineering from Unical in 2000. His research interests include agent-based computing, wireless (body) sensor networks, and the Internet of Things. He is an author of over 400 papers in international journals, conferences, and books. He is (founding) Series Editor of the IEEE Press Book Series on Human-Machine Systems and Editor-in-Chief of the Springer Internet of Things series, and an Associate Editor of many international journals including IEEE TAC, IEEE THMS, IEEE IoTJ, IEEE SJ, IEEE SMCM, Information Fusion, JNCA, EAAI, and so on. He is cofounder and CEO of SenSysCal S.r.l., a Unical spinoff focused on innovative IoT systems.